\documentclass[10pt,twocolumn,a4paper]{article}
\usepackage[utf8]{inputenc}
\usepackage{url}
\usepackage{graphicx}
\usepackage{authblk}

\usepackage{algorithm}
\usepackage{algorithmic}

\usepackage{amsmath}
\usepackage{soul}
\usepackage[hidelinks]{hyperref}
\usepackage{amsmath,amssymb}
\usepackage{booktabs, siunitx}
\usepackage{subcaption}
\usepackage{tablefootnote}
\usepackage{multirow}
\usepackage{todonotes}
\newcommand{\etal}{\textit{et al.}}

\usepackage{hyperref}

\usepackage{fancyhdr}
\title{Enforcing perceptual consistency on Generative Adversarial Networks by using the Normalised Laplacian Pyramid Distance}
\author[1]{Alexander Hepburn}
\author[2]{Valero Laparra}
\author[1]{Ryan McConville}
\author[1]{Raul Santos-Rodgriguez}

\affil[1]{Department of Engineering Mathematics, University of Bristol}
\affil[2]{Image Processing Lab, Universitat de Valencia}

\date{\vspace{-5ex}}

\begin{document}
\maketitle
\thispagestyle{fancy}

\begin{abstract}
In recent years there has been a growing interest in image generation through deep learning. While an important part of the evaluation of the generated images usually involves visual inspection, the inclusion of human perception as a factor in the training process is often overlooked. In this paper we propose an alternative perceptual regulariser for image-to-image translation using conditional generative adversarial networks (cGANs). To do so automatically (avoiding visual inspection), we use the Normalised Laplacian Pyramid Distance (NLPD) to measure the perceptual similarity between the generated image and the original image. The NLPD is based on the principle of normalising the value of coefficients with respect to a local estimate of mean energy at different scales and has already been successfully tested in different experiments involving human perception. We compare this regulariser with the originally proposed L1 distance and note that when using NLPD the generated images contain more realistic values for both local and global contrast. 
\end{abstract}

\section{Introduction}
Deep learning methods have become state-of-the-art in conditional and unconditional image generation~\cite{radford2015unsupervised} 
, achieving great success in numerous applications. 
Image-to-image translation is one such application, where the task involves the translation of one scene representation into another representation. 
Neural network architectures are able to generalise to different datasets and learn various translations between scene representations. For instance  
obtaining realistic scenes from segmented labels for training autonomous car system~\cite{pix2pix2016}.

Most state of the art methods in image-to-image translation typically use a Generative Adversarial Network (GAN) loss with regularisation.
This regularisation is typically with functions such as the L1, L2 or mean squared error (MSE)~\cite{pix2pix2016}. However, these do not account for the human visual system's perception of quality. For example, the L1 loss uses a pixel to pixel similarity which fails to capture the global or local structure of the image. 

The main objective of these methods is to generate images that look \emph{perceptually} indistinguishable from the training data to humans.
Despite this, metrics which attempt to capture different aspects of images that are important to humans are ignored.
Although neural networks seem to transform the data to a domain where the Euclidean distance induce a spatially invariant image similarity metric, given a diverse enough training dataset~\cite{zhang2018unreasonable}, we believe that explicitly including key attributes of human perception is an important step when designing similarity metrics for image generation.

Here we propose the use of a perceptual distance measure based on the human visual system that encapsulates the structure of the image at various scales, whilst normalising locally the energy of the image; the Normalised Laplacian Pyramid Distance (NLPD). This distance was found to correlate with human perceptual quality when images are subjected to different perturbations 
~\cite{laparra2016}. 
The main contributions of this paper are as follows; we argue that human perception should be used in the objective function of cGANs and propose a regulariser that measures human perceptual quality called NLPD; we evaluate the method comparing with L1 loss regularisation using no-reference image quality metrics, image segmentation accuracy and Amazon Mechanical Turk survey; we show improved performance over L1 regularisation, demonstrating the benefits of image quality metric inspired by the human perceptual system in the objective function.

\section{Related Work}
Image-to-image translation systems designed by experts can only be applied to their respective representations, unable to learn different translations~\cite{hertzmann2001image}. Neural networks are able to generalise and learn a variety of mappings and have proven to be successful in image generation~\cite{radford2015unsupervised}.

Generative Adversarial Networks (GANs) aim to generate data indistinguishable from the training data~\cite{goodfellow2014generative}. The generator network $G$ learns a mapping from noise vector $z$ to target data $y$, $G(z) \xrightarrow{} y$ and the discriminator network $D$ learns mapping from data $x$ to label $[0, 1]$, $D(x) \xrightarrow{} [0, 1]$ corresponding to whether the data is real or generated. GANs have become successful in complex tasks such as image generation~\cite{radford2015unsupervised}.
Conditional GANs (cGANs) learn a generative model that will sample data according to some attribute e.g. `generate data from class A'~\cite{pix2pix2016}. 

An application of cGANs is image-to-image translation, where the generator is conditioned on an image to generate a corresponding output image. In ~\cite{pix2pix2016} the cGAN objective function has a structured loss, the GAN considers the structure of the space and pixels are conditionally-dependent on all other pixels in the image. 

Optimising for the GAN objective alone creates images that lack outlines for the objects in the semantic label map and a common practice is to use the L2 or L1 loss as a reconstruction loss. 
Isola \etal{} preferred the L1 loss, finding that the L2 loss encouraged smoothing in the generated images. The L1 loss is a pixel level metric, meaning it only considers the distance between single pixel values ignoring the local structure that could capture perceptual similarity.



When the output of a algorithm will be evaluated by human observers, the image quality metric (IQM) used in the optimisation objective should take into account human perception.

In the deep learning community, the VGG loss~\cite{dosovitskiy2016generating} has been used to address the issue of generating images using perceptual similarity metrics. This method relies on using a network trained to predict perceptual similarity between two images. It has been shown to be robust to small structural perturbations, such as rotations, which is a downfall of more traditional image quality metrics such as the structural similarity index (SSIM). 
However, the architecture design and the optimisation takes no inspiration from human perceptual systems and treats the problem as a simple regression task; given image A and image B, output a similarity that mimics the human perceptual score.


There is a long tradition of IQMs based on human perception. The most well know is the SSIM or its multi scale version (MS-SSIM)~\cite{wang2003multiscale}. These distances focus on predicting the human perceptual similarity, but their formulation is disconnected from the processing pipeline followed by the human visual system. On the contrary, metrics like the one proposed by Laparra \etal{} are inspired by the early stages of the human visual cortex and show better performance in mimicking human perception than SSIM and MS-SSIM in different human rated databases. In this work we use this metric, the Normalised Laplacian Pyramid Distance (NLPD), proposed by Laparra \etal{}~\cite{laparra2016}.



\section{NLPD}
The Laplacian Pyramid is a well known image processing algorithm for image compression and encoding~\cite{burt1983laplacian}. The image is encoded by performing convolutions with a low-pass filter and then subtracting this from the original image multiple times, each time downsampling the image. The resulting filtered versions of the image have low variance and entropy and as such can be expressed with less storing information.

Normalised Laplacian Pyramid (NLP) extends the Laplacian pyramid with a local normalisation step on the output of each stage. These two steps are similar to the early stages of the human visual system. Laparra \etal{} proposed an IQM based on computing distances in the NLP transformed domain, the NLPD~\cite{laparra2016}. It has been shown that NLPD correlates better with human perception than the previously proposed IQMs. NLPD has been employed successfully to optimise image processing algorithms, for instance 
to perceptually optimised image rendering processes~\cite{laparra2017perceptually}. It has also been shown that the NLP reduces the correlation and mutual information between the image coefficients, which is in agreement with the efficient coding hypothesis~\cite{Barlow}, proposed as a principle followed by the human brain. 

Specifically NLPD uses a series of low-pass filters, downsampling and local energy normalisation to transform the image into a `perceptual space'. 
A distance is then computed between two images within this space. The normalisation step divides by a local estimate of the amplitude. The local amplitude is a weighted sum of neighbouring pixels where the weights are pre-computed by optimising a prediction of the local amplitude using undistorted images from a different dataset. The downsampling and normalisation are done at $N$ stages, a parameter set by the user. An overview of the architecture is detailed in~\cite{laparra2016}.

After computing each $y^{(k)}$ output at every stage of the pyramid, the final distance is the root mean square error between the outputs of two images:
\begin{equation}
\mathcal{L}_{NLPD} = \frac{1}{N} \sum^N_{k=1}\frac{1}{N_s^{(k)}}||y_1^{(k)} - y_2^{(k)}||_2,
\end{equation} where $k$ defines the stage, $N$ is the number of stages in the pyramid, $N_s^{(k)}$ is the square root of number of pixels at scale $k$, and $y_1^{(k)}$ and $y_2^{(k)}$ are the outputs for the training and the generated images respectively.

Qualitatively, the transformation to the perceptual space defined by NLPD transforms images such that the local contrast is normalised by the contrast of each pixels neighbours. This leads to NLPD heavily penalising differences in local contrast. Using NLPD as a regulariser enforces a more realistic local contrast and, due to NLPD observing multiple resolutions of the image, it also improves global contrast

In image generation, perceptual similarity is the overall goal; fooling a human into thinking a generated image is real. As such, NLPD would be an ideal candidate regulariser for generative models, GANs in particular.




\subsection{NLPD as a Regulariser}
For cGANs, the objective function is given by 
\begin{align}
   \mathcal{L}_{cGAN}(G,D) = &\mathbb{E}_{x, y}[\log D(x, y)]+ \\ &\mathbb{E}_{x, z}[\log (1-D(G(x,z))] \nonumber
\end{align}
where $G$ maps image $x$ and noise $z$ to target image $y$, $G: {x, z} \xrightarrow{} y$ and $D$ maps image $x$ and target image $y$ to a label in $[0, 1]$. 
With the L1 regulariser proposed by Isola \etal{}~\cite{pix2pix2016} for image-to-image translation, this becomes 
 \begin{equation}\label{eqn:cganl}
     \mathcal{L}_{cGAN}(G,D) + \lambda \mathcal{L}_{L1},
 \end{equation} 
where $\mathcal{L}_{L1} = \mathbb{E}_{x,y,z}[||y-G(x,z)||_1]$ and $\lambda$ is a tunable hyperparameter. 

In this paper we propose replacing the L1 regulariser $\mathcal{L}_{L1}$ with a NLPD regulariser. In doing so the entire objective function is given by
\begin{equation}\label{eqn:cganNLPD}
    \mathcal{L}_{cGAN}(G,D) + \lambda \mathcal{L}_{NLPD}.
\end{equation}

In the remainder of the paper Eq.~\eqref{eqn:cganl} will be denoted by cGAN+L1 and Eq.~\eqref{eqn:cganNLPD} by cGAN+NLPD.

NLPD involves $3$ convolutions per stage in the pyramid, with the same filter convoluted to each colour channel of the input. This is more computationally expensive than $L1$ loss, but relative to the training procedure of training a GAN, the increase in time is negligible. Using computational packages, the process of transforming images into the perceptual space can be appended to the  computation graph as extra convolutional layers. 

\section{Experiments}

We evaluated our method on three public datasets; the  Facades dataset~\cite{tylevcek2013spatial}, the Cityscapes dataset~\cite{cordts2016cityscapes} and a Maps dataset~\cite{pix2pix2016}.  Colour images were generated from semantic label maps for the Facades and the Cityscapes datasets. The Facades dataset is a set of architectural label drawings and the corresponding colour image for various buildings. The Cityscapes dataset is a collection of label maps and colour images taken from the a front facing car camera. For the Cityscapes dataset, images were resized to a resolution of $256\times256$ and after generating the images they were resized to the original dataset aspect ratio of $512\times256$.
The third dataset is the Maps dataset constructed by Isola \etal{}. It contains a map layout image of an area taken from Google Maps and the corresponding aerial image resized to a resolution of $256\times256$. 

The objective of all of these tasks is to generate a RGB image from the textureless label map. For all datasets, the same train and test splits were used as in the pix2pix paper, in order to ensure a fair comparison.

\subsection{Experimental Setup}
For all experiments, the architecture of the generator and discriminator is the same as in ~\cite{pix2pix2016}. The generator is a U-net with skip connections between each mirroring layer. The discriminator observes $70\times70$ pixel patches at a time, with dropout applied at training. Full architecture can be found in the paper by Isola \etal{}.
As in~\cite{pix2pix2016} we used the Adam optimiser 
with learning rate $0.0002$, trained each network for $200$ epochs, and used batch-size of 1 with batch normalisation (equivalent to use instance normalization). 
Random cropping and mirroring were applied during training.

We used $\lambda =100$ for the L1 regulariser, the value used by Isola \etal{}~\cite{pix2pix2016}. We set $\lambda=15$ for NLPD so that both loss terms have a similar order of magnitude.
The number of stages was chosen as $N=6$. 
The normalisation filters were found by optimising to recover the original local amplitude from perturbed images in the McGill dataset~\cite{olmos2004biologically}. These weights were found using monochromatic images so the normalisation is applied to each channel independently. 

\subsection{Evaluation}

Evaluating generative models is a difficult task~\cite{theis2015note}. Therefore we have performed different experiments to illustrate the improvement in the performance when using NLPD as regulariser. 
In image-to-image translation, there is information in the form of the label map that images were generated with. 
A common metric is evaluating the performance of a network trained on the ground truth at image segmentation using generated images~\cite{pix2pix2016,wang2018high}. Generated images which achieve higher performance can be considered more realistic. One architecture that has been used for image segmentation is the fully convolution network (FCN)~\cite{long2015fully}.

\begin{figure*}[htb]
    \centering
    \begin{subfigure}[c]{0.94\textwidth}
        \centering
        \includegraphics[width=\textwidth]{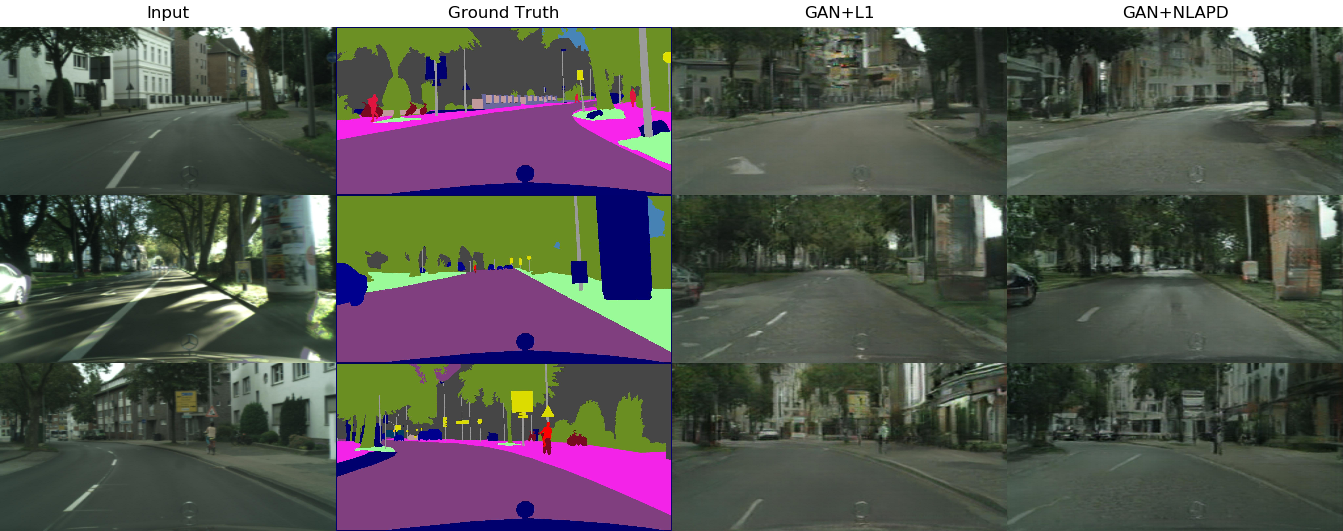}
    \end{subfigure}
    \caption{Images generated from label maps taken from the {Cityscapes} validation set. Images were generated at a resolution of $256\times256$ and then resized to the original aspect ratio of $512\times256$.}
    \label{fig:cityscapes256}
\end{figure*}

\begin{figure*}[htb]
    \centering
    \begin{subfigure}[b]{0.48\textwidth}
        \centering
        \includegraphics[width=\textwidth]{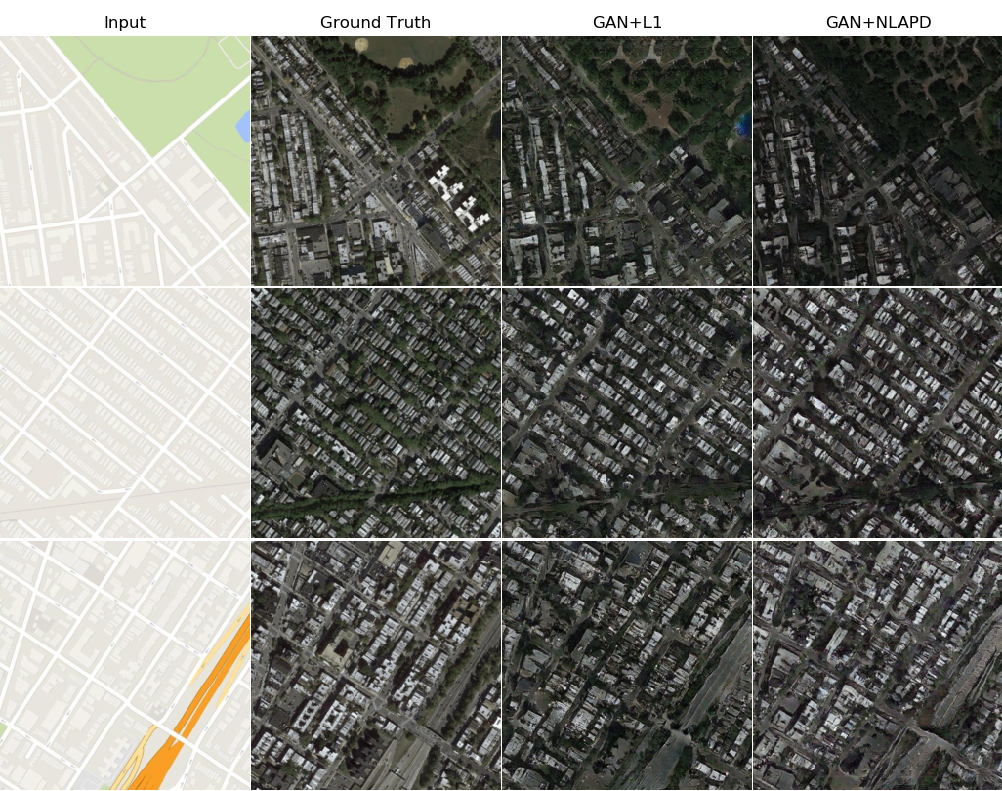}
        \caption{{Maps}}
        \label{fig:maps}
    \end{subfigure}
    \begin{subfigure}[b]{0.48\textwidth}
        \centering
        \includegraphics[width=\textwidth]{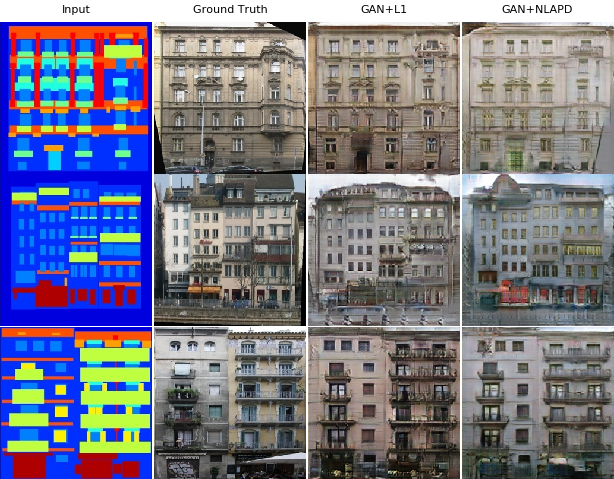}
        \caption{{Facades}}
        \label{fig:facades}
    \end{subfigure}
    \caption{Images generated from the (a) {Maps} and (b) {Facades} datasets at a resolution of $256\times256$.}
    \label{fig:mapsfacades}
\end{figure*}


As in~\cite{pix2pix2016} we trained a image segmentation network on the Cityscapes dataset and evaluated it on generated images.
3 image segmentation metrics are calculated. Per-pixel accuracy is the percentage of pixels correctly classified, per-class accuracy is the mean of the accuracies for classes and class IOU is the intersection over union, the percentage overlap between the ground truth and the predicted label map.



When generating an image from a label map, the ground truth is just one possible solution and there exists many feasible solutions. 
We include two no-reference image quality metrics to more thoroughly evaluate the generated images, namely BRISQUE~\cite{mittal2012no} and NIQE~\cite{mittal2013making} which judge how natural an image appears.

Our objective is to generate images which look perceptually similar to the original images. We have performed an experiment using Amazon Mechanical Turk (AMT) asking humans to judge ``Which image looks more natural?''.
A random subset of $100$ images were chosen from the validation set of each dataset and $5$ unique decisions were gathered per image. 

\subsection{Results}
Results of images generated, from the test set, using the proposed procedure and the L1 baseline for the three datasets are presented in Figs.~\ref{fig:cityscapes256}, \ref{fig:maps}, and \ref{fig:facades}. Fig.~\ref{fig:maps} shows aerial images generated from a map: NLPD produces realistic textures, whereas L1 has repeating patterns. In the {Cityscapes} dataset the contrast appears more realistic, e.g., the white in the sky is lighter in Fig.~\ref{fig:cityscapes256}, which could result in users preferring these images. In images generated from the {Facades} dataset, it is hard to visually find differences between the methods (Fig.~\ref{fig:facades}).

Table \ref{table:fcn-scores-losses} shows the FCN-scores for the images generated using the Cityscapes database. The NLPD images show improvement over the L1 regularisation in the per-pixel accuracy and class IOU. This infers that the NLPD images contain more features of the original dataset according to the FCN network.  The ground truth accuracy less than $1$ because the segmentation network is trained on images of resolution $256\times256$, then resized to the resolution of the label map, $2048\times1024$.


\begin{table}[htb]
\centering
\small
\begin{tabular}{@{}llll@{}}
\toprule
\multirow{2}{*}{\begin{tabular}[c]{@{}l@{}}Loss\\ Function\end{tabular}} & \multicolumn{3}{l}{BRISQUE (NIQE) Scores} \\ \cmidrule(l){2-4} 
 & Facades & \begin{tabular}[c]{@{}l@{}}Cityscapes\end{tabular} & Maps \\ \midrule
cGAN+L1 & 30.1 (5.2) & 26.6 (3.9) & 30.6 (4.7) \\
cGAN+NLPD & 30.1 (5.2) & 24.5 (3.6) & 29.0 (4.6) \\ \midrule
Ground Truth & 37.3 (7.3) & 25.4 (3.1) & 28.5 (3.4) \\ \bottomrule
\end{tabular}
\caption{BRISQUE and NIQE scores. The lower the score, the more natural the image is.}
\label{table:bn}
\end{table}

Table \ref{table:bn} shows the scores for both the BRISQUE and NIQE image quality metrics. 
A lower value means a more natural image. On average, NLPD regularisation achieves lower values in both metrics. For Cityscapes and Maps, NLPD is close to the ground truth scores. The ground truth scores for the Facades dataset are worse than the generated images due to the triangles that are in the Facades training set, to crop neighbouring buildings. 

For AMT experiments, the percentage of users that found the NLPD images more natural was above chance for {Maps} ($52.37\%$) and {Cityscapes} ($56.16\%$), and similar for Facades ($50.04\%$).
\begin{table}[htb]
\centering
\small
\begin{tabular}{@{}llll@{}}
\toprule
{Loss} & {\begin{tabular}[c]{@{}l@{}}Per-Pixel\\ Accuracy\end{tabular}} & {\begin{tabular}[c]{@{}l@{}}Per-Class\\ Accuracy\end{tabular}} & {\begin{tabular}[c]{@{}l@{}}Class\\IOU\end{tabular}}\\ \midrule
{cGAN+L1} & $0.71\pm0.15$ & $0.25\pm0.05$ & $0.18\pm0.04$\\
{cGAN+NLPD} & $0.74 \pm 0.09$ & $0.25\pm 0.04$ & $0.19\pm 0.04$ \\ \midrule
{Ground Truth} & $0.80\pm 0.09$ & $0.26\pm 0.04$ & $0.21\pm 0.04$ \\
\bottomrule
\end{tabular}
\caption{FCN-scores  for Cityscapes dataset. Reported is the mean and standard deviation across the test set.} 
\label{table:fcn-scores-losses}
\end{table}



\section{Conclusion}
Taking into account human perception in machine learning algorithms is challenging and usually overlooked in image generation. We detailed a procedure to take into account human perception in a cGAN framework. We propose to modify the standard objective by incorporating a term that accounts for perceptual quality by using the NLPD. We illustrate its behaviour in the image-to-image translation for a variety of datasets. The suggested objective shows better performance in all the evaluation procedures. It also has a better segmentation accuracy using a network trained on the original dataset, and produces more natural images according to two no-reference image quality metrics. In human experiments, users preferred for the images generated using our proposal over those generated using L1 regularisation. 

\bibliographystyle{abbrv}
\small
\bibliography{ref}

\end{document}